%% file: root.tex
\documentclass[letterpaper, 10 pt, conference]{ieeeconf}  

\IEEEoverridecommandlockouts                              

\usepackage{epsfig} 
\usepackage{mathptmx} 
\usepackage{times} 
\usepackage{amsmath} 
\usepackage{amssymb}  
\usepackage{amsthm}
\usepackage{color}
\usepackage{xcolor}
\usepackage{preambles}
\usepackage{used_variables}
\usepackage{amsmath}
\usepackage{graphicx}
\usepackage{url} 
\usepackage[format=plain,labelsep=period]{caption}
\usepackage{wrapfig}
\usepackage[list=on,listformat=simple]{subcaption}
\usepackage{hyperref}
\usepackage{url}
\usepackage{caption}
\usepackage{subcaption}
\usepackage{empheq}
\usepackage[most]{tcolorbox}

\hypersetup{colorlinks = true, citecolor=black}

\DeclareMathAlphabet{\mathcal}{OMS}{cmsy}{m}{n}
\DeclareMathOperator*{\argminB}{argmin}   

\title{\LARGE \bf
Safe Whole-Body Task Space Control for Humanoid Robots}
\author{Victor C. Paredes$^{1}$ and Ayonga Hereid$^{1}$
\thanks{*This work was supported in part by the National Science Foundation under grant FRR-21441568. }%
\thanks{$^{1}$Mechanical and Aerospace Engineering, Ohio State University, Columbus, OH, USA. {\tt\footnotesize (paredescauna.1, hereid.1)@osu.edu.}}%
}

\usepackage[capitalise]{cleveref}

\usepackage[noadjust]{cite}

\newtcbox{\mymath}[1][]{%
    nobeforeafter, math upper, tcbox raise base,
    enhanced, colframe=blue!30!black,
    colback=blue!30, boxrule=1pt}

\begin{document}

\maketitle
\thispagestyle{empty}
\pagestyle{empty}

\begin{abstract}
Complex robotic systems require whole-body controllers to deal with contact interactions, handle closed kinematic chains, and track task-space control objectives. However, for many applications, safety-critical controllers are important to steer away from undesired robot configurations to prevent unsafe behaviors.
A prime example is legged robotics, where we can have tasks such as balance control, regulation of torso orientation, and, most importantly, walking. As the coordination of multi-body systems is non-trivial, following a combination of those tasks might lead to configurations that are deemed dangerous, such as stepping on its support foot during walking, leaning the torso excessively, or producing excessive centroidal momentum, resulting in non-human-like walking. 
To address these challenges, we propose a formulation of an inverse dynamics control enhanced with exponential control barrier functions for robotic systems with numerous degrees of freedom. Our approach utilizes a quadratic program that respects closed kinematic chains, minimizes the control objectives, and imposes desired constraints on the Zero Moment Point, friction cone, and torque. More importantly, it also ensures the forward invariance of a general user-defined high Relative-Degree safety set.
We demonstrate the effectiveness of our method by applying it to the 3D biped robot Digit, both in simulation and with hardware experiments.
\end{abstract}

\input{sections/introduction}
\input{sections/robot_dynamics}

\input{sections/task_space_control}

\input{sections/control_bf}

\input{sections/results}

\input{sections/conclusions}

\bibliographystyle{IEEEtran}
\bibliography{references.bib}

\end{document}

%% file: sections/introduction.tex
\section{Introduction}
Humanoid robots have emerged as a highly promising platform for performing complex tasks in human-centered environments due to their anthropomorphic structure. With dedicated legs and arms, these robots are well-equipped to simultaneously walk and manipulate objects. However, effectively coordinating the movements of legs and arms in a safe and stable manner is not a straightforward task. The dynamic coupling between these components makes independent control prone to instability and subpar performance. Therefore, the development of a holistic controller that can safely coordinate the entire body is necessary to successfully accomplish these tasks while respecting the robot's dynamics. Safety considerations are of utmost importance when deploying complex robots in real-world scenarios. Even if a desired task can be controlled, poor performance may result if safety measures are not adequately taken into account. For instance, a non-safe task-space controller may successfully track a desired swing foot pose of a humanoid robot, but it may fail to check if any leg joints are approaching mechanical limits or if the robot is at risk of self-collision. By incorporating a safety layer, the controller explicitly verifies and enforces control solutions that prioritize safety.

\begin{figure}[t]
\vspace{2mm}
\centering
\includegraphics[width=0.9\columnwidth]{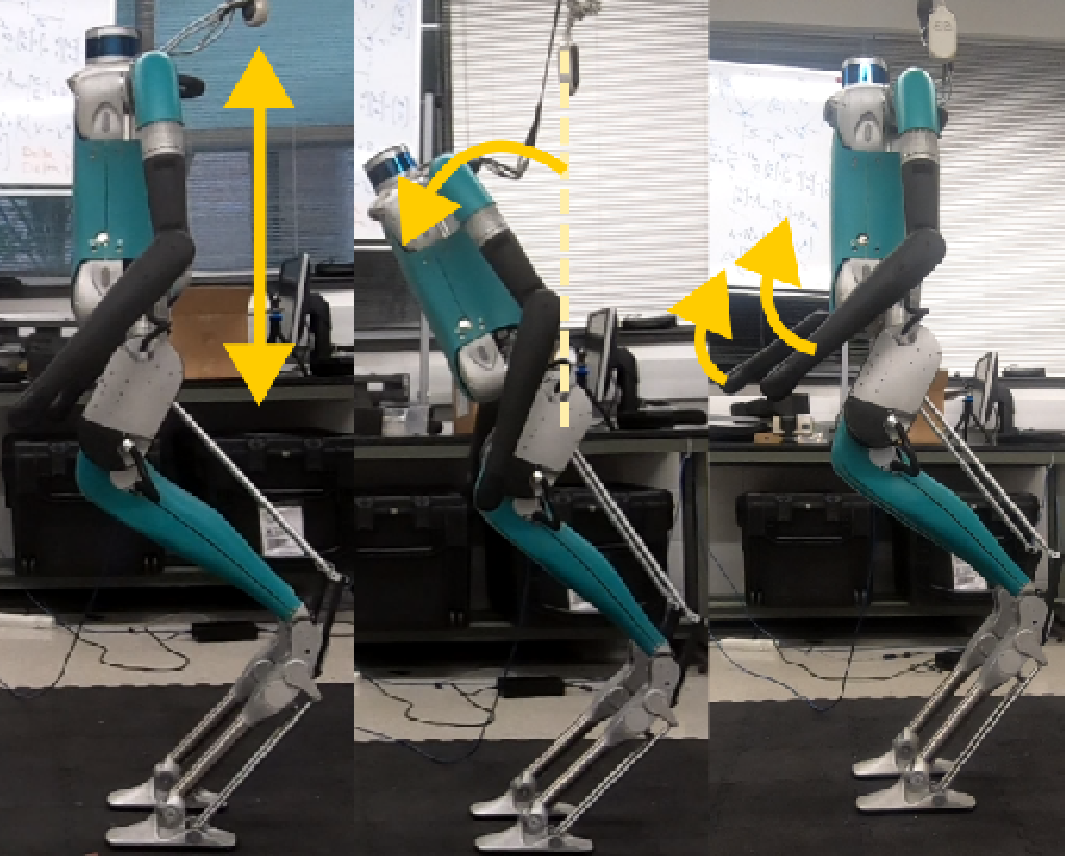}
\caption{The humanoid robot Digit performs different tasks involving CoM motion, torso orientation, and arm movements using our proposed safe whole body task space control.}
\vspace{-5mm}
\label{fig:digit}
\end{figure}

Safety-critical control systems has been extensively studied using barrier certificates \cite{prajna2006barrier, prajna2004safety}, which are defined by state-dependent sets described by a function that must remain positive. By appropriately restricting the controller action, a control algorithm can ensure that this safe set remains invariant. Recent research has introduced the concept of Control Barrier Functions (CBFs) within the context of Lyapunov Theory \cite{hsu2015control} and optimization \cite{nguyen2016exponential, ames2019control} to express safety certificates. Notably, the formulation of Exponential Control Barrier Functions (ECBF) has been a significant advancement, as it allows safety constraints to have arbitrarily high relative degrees, making it less restrictive for real-world applications as it enables to go beyond geometrical constraints~\cite{del2017joint, khazoom2022humanoid}. Leveraging a non-linear controller within an optimization framework becomes advantageous when incorporating ECBFs, as it enables the restructuring of the controller based on recent developments in task-space control.
Task-space control has been extensively studied using both model-based inverse dynamics \cite{nakanishi2008operational} and model-free inverse kinematics approaches \cite{paredes2022resolved}. The use of inverse dynamics offers the advantage of considering model constraints such as contact constraints, friction cone, zero moment point, and torque limits. However, controlling bipedal systems presents challenges due to their intrinsic under-actuation and floating base. Nakanishi et al. \cite{nakanishi2007inverse} proposed a closed-form control solution that relies on estimating contact forces to obtain the constrained dynamics and the Jacobian projection of the task space objectives. An improved version of this controller \cite{mistry2010inverse} utilizes orthogonal decomposition to work in a reduced dimensional space and avoids the need for estimating contact forces. However, these formulations do not explicitly incorporate contact wrenches, limiting their ability to introduce additional relevant constraints such as the zero moment point or friction.

To effectively address these constraints, it is crucial to explicitly consider the contact wrenches in an inverse dynamics controller that leverages the dynamics of user-defined general task outputs. Herzog et al. \cite{herzog2014balancing} proposed a Quadratic-Programming (QP) formulation that incorporates the robot dynamics and treats the contact wrenches as decision variables. This QP-based approach offers advantages in terms of reducing the complexity of matrix operations and enabling the handling of multiple constraints. Building upon this foundation, Reher et al. \cite{reher2020inverse} introduced a similar QP optimization structure to construct a Control Lyapunov Function (CLF) that respects constraints such as the zero moment point (ZMP), contact, and friction cone. Their formulation utilized acceleration, torque, and constraint wrenches as optimization variables, providing enhanced control capabilities over these variables. However, the inverse dynamics formulations in these works do not consider safety.
The work of \cite{khazoom2022humanoid} provides a formulation of whole-body control with a Control Barrier Function (CBF) specifically designed for position-based objectives. 
On the other hand, Nguyen et al. \cite{nguyen20163d, nguyen2015safety} presented a QP-based controller that incorporates an Exponential Control Barrier Function (ECBF), a formulation that extends the CBF by the consideration of general state-space-based safety sets with arbitrary relative-degree. They employed a CLF-based controller and explicitly constructed an ECBF to constrain footstep placements on stepping stones. However, their optimization formulation utilized motor torques as decision variables, resulting in increased numerical complexity due to the inversion of the mass matrix. Moreover, the formulation does not consider friction or ZMP constraints important for more realistic implementations. To overcome this challenge, we extend the work of Reher et al. \cite{reher2020inverse} by reformulating the barrier functions as an acceleration-based exponential control barrier function (A-ECBF). This novel formulation avoids explicit dependence on torques and effectively alleviates the numerical cost associated with the QP formulation.

The main contribution of this paper is the development of a novel Quadratic-Programming (QP)-based safe inverse dynamics controller that offers several key advantages:
\begin{enumerate}
\item \textbf{Avoidance of mass matrix inversions:} Building upon previous work on inverse dynamic formulations, we leverage a numerically efficient program that explicitly considers joint accelerations, torques, and wrenches as decision variables. This approach eliminates the need for computationally expensive mass matrix inversions.
\item \textbf{Handling of kinematics constraints:} Our QP-based inverse dynamics formulation enables the straightforward incorporation of closed-loop kinematics and other essential constraints, such as contacts, zero moment point (ZMP), and the friction cone. This flexibility allows for more accurate and realistic modeling of the robot's behavior.
\item \textbf{Enforcement of safety through acceleration-based exponential control barrier functions (A-ECBFs):} To ensure the safety of the system, we construct an exponential control barrier function that guarantees the invariant behavior of a predefined safe set. By formulating this safety certificate as an inequality constraint dependent solely on joint accelerations, we exploit the inherent structure of the controller formulation.
\item \textbf{Application of safe control actions to a 3D humanoid robot:} We demonstrate the performance and effectiveness of our controller through extensive simulation and hardware experiments on a 3D humanoid robot. These experiments validate the controller's ability to achieve desired tasks while maintaining safety.
\end{enumerate}

The remainder of the paper is organized as follows. \secref{sec:dynamics} presents the mathematical modeling of humanoid robots with floating base coordinates, contacts, closed-loop mechanisms, and critical dynamics constraints. In \secref{sec:CBF-QP}, we present a task-space inverse dynamics control algorithm that is expressed as a quadratic program, followed by formulating an Acceleration-based Exponential Control Barrier Function that can be naturally included in the inverse dynamics formulation. In \secref{sec:results}, we showcase the effectiveness of our whole body controller and its safety enforcement with various tasks for the 3D bipedal robot, Digit (Fig. \ref{fig:digit}). This section provides empirical evidence of the controller's performance and demonstrates its ability to handle complex tasks.

\label{sec:intro}

%% file: sections/robot_dynamics.tex
\begin{figure}[t]
\centering
\includegraphics[width=\columnwidth]{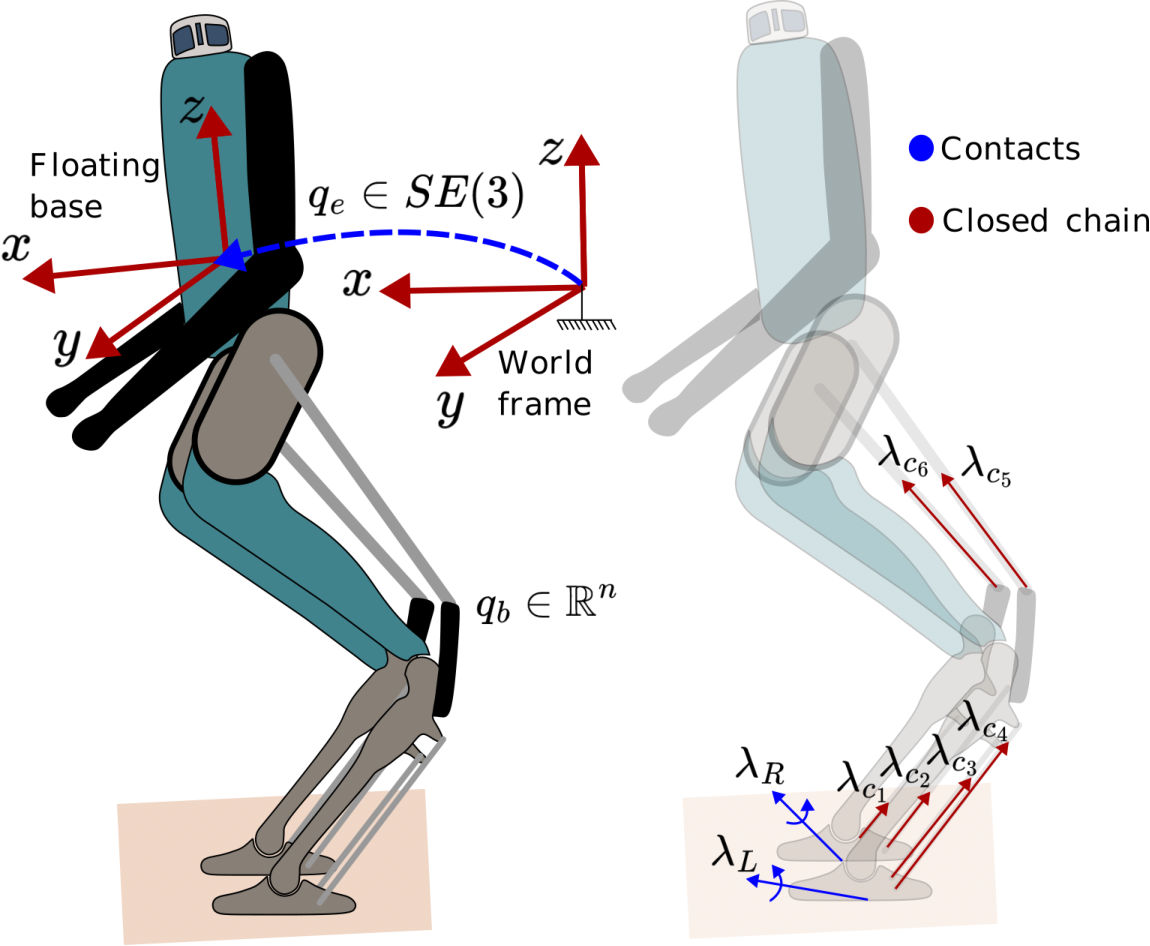}
\caption{A humanoid robot with floating base is described by its internal body coordinates $q_b$ and its floating coordinates $q_e$ as shown in the left. Furthermore, this robot experiences contact wrenches and forces due to the closed chain mechanisms as seen in the right.}
\label{fig:Digit_floatingBase}
\vspace{-4mm}
\end{figure}

\section{Humanoid Dynamics with Constraints}
\label{sec:dynamics}
The kinematics of humanoid robots can be described by a floating frame fixed to a base point, introducing respective floating coordinates $\qevec \in SE(3)$, as shown in \figref{fig:Digit_floatingBase}, and the body coordinates describe the relative motion of its joints. For a robot with $n_b$ joints, we represent body coordinates as $\qbvec \in \mathbb{R}^{n_b}$.
The configuration space $Q$ of a legged robot with a floating base, such as Digit as shown in \figref{fig:Digit_floatingBase}, then can be represented by $\qvec = [\qevec^\top, \qbvec^\top]^\top\in Q = \R^{n}$ with $n=n_b + 6$ being the total degrees of freedom of the robot. The dynamics of the multi-body system can be described by the Euler Lagrangian equations of motion~\cite{grizzle2014models}:
\begin{align}
    M(\qvec) \ddqvec + C(\qvec, \dqvec) \dqvec + G(\qvec) & = B \uvec + J(\qvec)^\top \lm,
    \label{eq: motion} 
\end{align}
where $M(\qvec) \in \mathbb{R}^{n \times n}$, $C(\qvec, \dqvec) \in \mathbb{R}^{n \times n}$, and $G(\qvec) \in \mathbb{R}^n$ are inertia matrix, Coriolis matrix, and gravity vector, respectively, $B \in \mathbb{R}^{n\times m}$ is the torque distribution matrix that maps the torque of the $m$ actuators $\uvec \in U \subset \mathbb{R}^{m}$, $\lm \in \mathbb{R}^{n_{h}}$ is the collection of constraint wrenches or external forces, and $J(\qvec) \in \mathbb{R}^{n_h \times n}$ its respective Jacobian matrix. Constraint wrenches are due to kinematic closed-chains, or contacts, as illustrated in \figref{fig:Digit_floatingBase} and discussed below.


 

\subsection{Closed kinematic chain constraints}
\begin{figure}[t]
\centering
\vspace{2.5mm}
\includegraphics[width=\columnwidth]{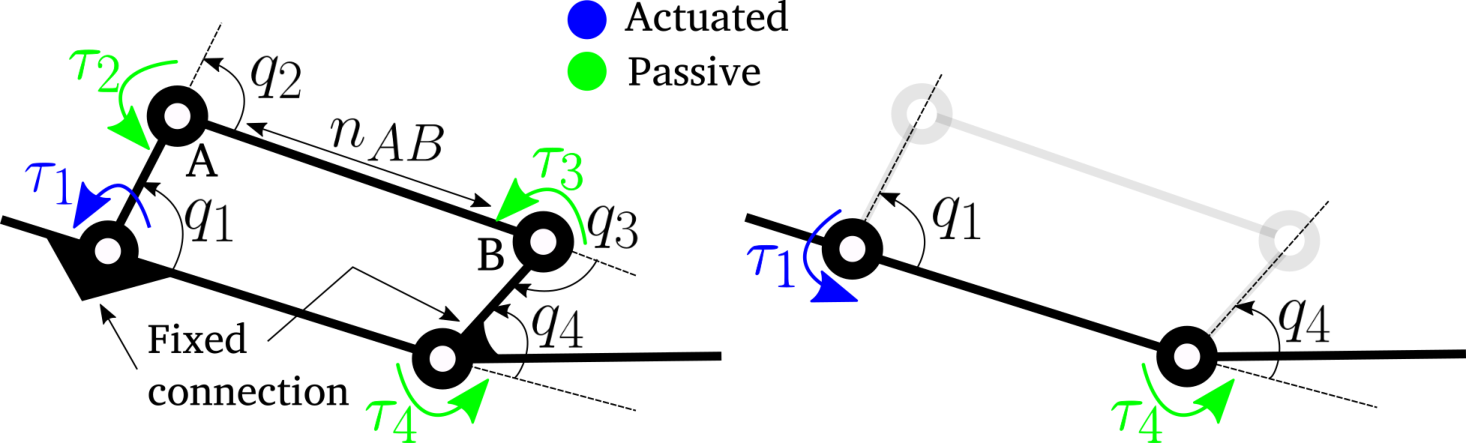}
\caption{The closed loop kinematics on the right can be expressed as the open kinematic chain on the left with a holonomic constraint that enforces $n_{AB}$ constant. This constraint will relate the actuated torque $\tau_1$ to the passive joint torque $\tau_4$.}
\label{fig:4barExample}
\vspace{-4mm}
\end{figure}

A closed kinematic chain, such as the four-bar linkage shown in \figref{fig:4barExample}, is popular in legged robot designs. However, many existing controllers do not explicitly address constraints associated with closed-chain kinematics~\cite{moro2018whole}. 
To model a closed kinematic chain without resolving the constrained dynamics, we can virtually disconnect each closing link and use their lengths given by $n_k(\qvec) \in \mathbb{R},  \forall k \in \Omega_{chain}$ (e.g., $n_{AB}$ in \figref{fig:4barExample}) to construct appropriate holonomic constraints that enforce $\ddot{n}_k(\qvec, \dqvec, \ddqvec) = 0, \forall k \in \Omega_{chain}$~\cite{reher2021control}. Note that typically the inertial effects of the connecting rods are neglected as they have a much smaller mass with respect to other links. The collection of such constraints generates,
\begin{align}
    J_{chain}(\qvec) \ddqvec + \dot{J}_{chain}(\qvec, \dqvec) \dot {\qvec} = 0
    \label{eq:holonomic}
\end{align}
where, $J_{chain}(\qvec)$ is the jacobian of the collection of the closed kinematic constraints. 

\subsection{Contact constraints}
The contact constraints are included whenever the robot must preserve contact with a point or a surface. 
While such ground contacts are unilateral, they can be modeled as holonomic constraints with additional inequality constraints describing the limitations imposed by friction and the direction of the normal force.
Similarly to the closed kinematic constraints, for a set of contacts $c \in \Omega_{cont}$, and their respective pose $n_c(\qvec) \in \mathbb{R}^6$, we enforce its invariance through:
\begin{align}
 J_{cont}(\qvec) \ddqvec + \dot{J}_{cont}(\qvec, \dqvec) \dot{\qvec} = 0
 \label{eq:contact}
\end{align}
Additionally, to impose friction constraints and avoid slipping we use the pyramidal friction cone approximation~\cite{grizzle2014models}. For each contact wrench we can decompose it into forces and moments as $\vec{\lambda}^c = \begin{bmatrix}  \lambda^c_{fx}, \lambda^c_{fy}, \lambda^c_{fz}, \lambda^c_{mx},\lambda^c_{my}, \lambda^c_{mz} \end{bmatrix}^T$ indexed by $c \in \Omega_{cont}$. We impose the following constraints that can be expressed in linear form,
\begin{align}
    |\lambda^c_{fx}| &\leq \frac{\mu}{\sqrt{2}} \lambda^c_{fz}, \\
    |\lambda^c_{fy}| &\leq \frac{\mu}{\sqrt{2}} \lambda^c_{fz}, \\
    \lambda^c_{fz} &> 0,
\end{align}
where $\mu$ is the friction coefficient between the feet and the ground, $\lambda^c_{fx}, \lambda^c_{fy}, \lambda^c_{fz}$ represent the linear force components of a contact wrench along $x,y,z$-axes of the contact frame. In some cases, to avoid rotational slipping, one must limit the normal moment by incorporating a soft-finger contact type:
\begin{align}
    |\lambda^c_{mz}| \leq \gamma \lambda^c_{fz}, 
\end{align}
where $\gamma$ is the torsional friction coefficient and $\lambda^c_{mz}$ is the rotational moment component of a wrench along the $z$-axis of the contact frame. 

\begin{figure}[t]
\centering
\vspace{2.5mm}
\includegraphics[width=\columnwidth]{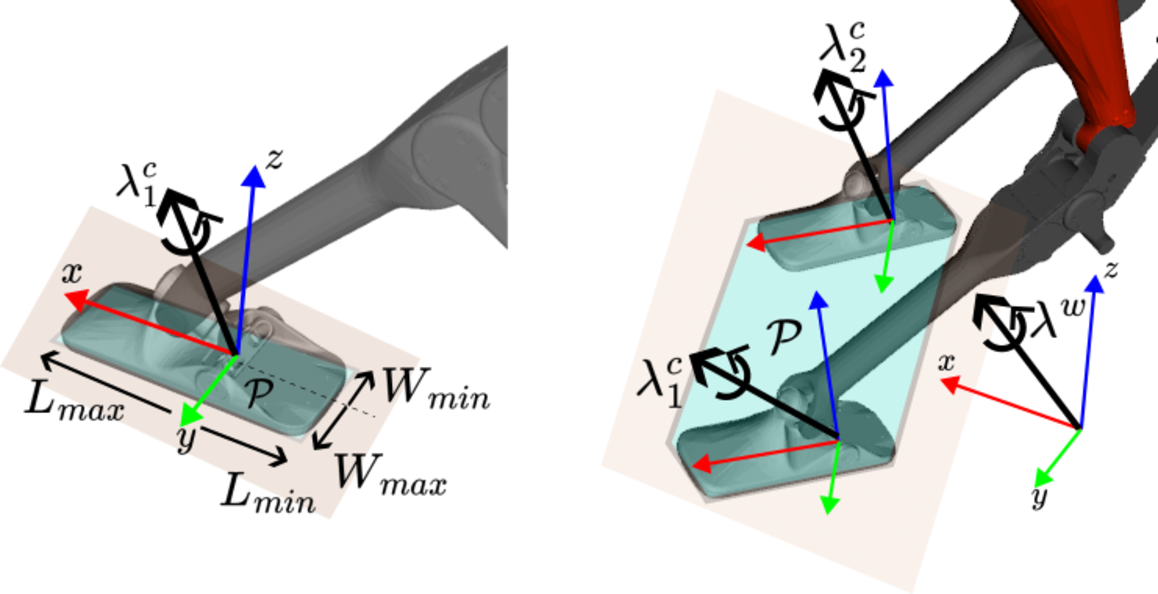}
\caption{Typical cases: One contact point on the left and two on the right; in both cases, the wrenches are projected into $\lambda^w$ for ZMP computations.}
\label{fig:ZMP}
\vspace{-4mm}
\end{figure}

Additional Zero Moment Point (ZMP) constraints should be enforced to prevent tipping over edges~\cite{vukobratovic2004zero, grizzle2014models}. The typical contact cases are single and double support configurations, as seen in Fig. \ref{fig:ZMP}. For a series of contacts, we need to project each wrench into a unique frame, for instance, the world frame with an associated global wrench $\lambda^w$. We can achieve this by applying the adjoint transformation (represented by $Ad_g$) to every single contact wrench to obtain the equivalent contact wrench expressed in the world frame:
\begin{align}
    \lambda^w = \sum_{i=1}^{N_c} Ad^T_{g_{c \rightarrow w}}(\qvec) \vec{\lambda}^c_i,
\end{align}
where $g_{c \rightarrow w}$ is the homogeneous transformation matrix of the world frame w.r.t. the contact frame $c \in \Omega_{cont}$.

The ZMP is computed as $p^x_{zmp} = \lambda^{w}_{my} /  \lambda^{w}_{fz}$, $p^y_{zmp} = -\lambda^{w}_{mx} / \lambda^{w}_{fz}$ and must be inside the support polygon $\mathit{P}$ defined by the contact geometry, as shown in \figref{fig:ZMP}. Thus, the ZMP constraint can be represented as:
\begin{align}
 \begin{bmatrix} p^x_{zmp} & p^y_{zmp} \end{bmatrix} \in \mathcal{P}.
\end{align}
where, $\mathcal{P}$ is the support region formed by the contact bodies as seen in Fig. \ref{fig:ZMP} for both single support and double support.  

%% file: sections/task_space_control.tex
\section{Safe Task Space Whole Body Control via CBF-QP}
\label{sec:CBF-QP}
This section presents an inverse dynamics controller based on quadratic programming (QP) that incorporates holonomic constraints arising from contacts and closed chain mechanisms. This approach avoids the need for matrix inversions and the calculation of constrained dynamics, providing numerical benefits.
Subsequently, we introduce our paper's main contribution, which is applying an Exponential Control Barrier Function (ECBF) to ensure the invariance of a safe set. Specifically, we formulate an Acceleration-based ECBF compatible with the whole-body inverse dynamics control as shown in Fig. \ref{fig:framework}.

\subsection{Task Space Inverse dynamics}
We consider task space outputs $y(t, \qvec) \in \mathbb{R}^m$ as positions and orientations of the robot's representative points or reference frames. These outputs can specify the robot's desired behavior, such as moving to a particular position or following a trajectory. Typical outputs may include the center of mass (CoM) position and orientation and the end-effector poses of the robot's arms and legs. For simplicity, we present an output that is of relative degree two, given as:
\begin{align}
    y(t, \qvec) &:= y^a(\qvec) - y^d(t),  
\end{align}
    where, $y^a(\qvec), y^d(t) \in \mathbb{R}^m$ are the actual and desired outputs respectively. Further, the first and second order derivatives of outputs are:    
\begin{align}
    \dot{y}(t,\qvec,\dqvec) &= J_y(\qvec) \dqvec - \dot{y}^d(t), \\
    \ddot{y}(t,\qvec,\dqvec,\ddqvec)&= \dot{J}_y(\qvec,\dqvec) \dqvec + J_y(\qvec) \ddqvec - \ddot{y}^d(t)
    \label{eq:yddot}
\end{align}
where $J_y(\qvec) \in \mathbb{R}^{m \times n}$ is the jacobian of $y^a(\qvec)$. 

We formulate the inverse dynamics problem as an optimization program using the decision variable $X = [\ddot{q}^T, u^T, \lambda^T]^T$. Hence, the constrained robot dynamics in \eqref{eq: motion} - \eqref{eq:contact} can be rewritten as:
\begin{align}
    \underbrace{\begin{bmatrix} M(q) & -B & -J(q)^T \end{bmatrix}}_{D_{eq}(q)} X &= -C(q,\dot{q})\dot{q} - G(q),\\
    \underbrace{\begin{bmatrix} J(q) & 0 & 0 \end{bmatrix}}_{C_{eq}(q)} X &= - \dot{J}(q,\dot{q}) \dot{q}, 
\end{align}
where $J(q) = \begin{bmatrix} J_{chain}(q) \\  J_{cont}(q) \end{bmatrix}$.
To obtain stable output dynamics, we enforce a linear output dynamics $\ddot{y}(t,q,\dot{q},\ddot{q}) = -K_p y(t,q) - K_d \dot{y}(t,q,\dot{q})$, which holds true whenever,
\begin{align}
    J_y(q) \ddot{q} + \dot{J}_y(q,\dot{q}) \dot{q} = \underbrace{-K_p y(t,q) - K_d \dot{y}(t,q,\dot{q}) + \ddot{y}^d(t) }_{y^*}
\end{align}
where, $K_p, K_d \in \mathbb{R}^{m \times m}$ are positive definite gain matrices.
We formulate the QP-based inverse dynamics controller as,
\begin{align*}
    X^* = \argminB_{X \in \mathcal{X}} &|| J_y(q) \ddot{q} + \dot{J}_y(q, \dot{q}) \dot{q} - y^*||^2 +  X^T \Gamma X \\
    \st & D_{eq}(q) X = -C(q,\dot{q}) - G(q) \\
    & C_{eq}(q) X = -J(q) \dot{q} 
\end{align*}
where $\mathcal{X}$ captures the domain constraints (e.g., torque limits, friction cone, and ZMP constraints) expressed on the optimization variables. 
We use the regularization term $X^T \Gamma X$, with $\Gamma > 0$ with a weight represented by a diagonal matrix, to avoid high-frequency changes in the optimization variables. 

%% file: sections/control_bf.tex
\subsection{Exponential control barrier functions}
To enhance the capabilities of the inverse dynamics control, we introduce safety through the use of Exponential Control Barrier Functions (ECBF)~\cite{ames2019control}. They ensure the forward invariance of a user-defined safe set. Moreover, we will utilize an acceleration-based formulation of the ECBF (A-ECBF), which leverages the decision variable $X$ to maintain the inverse dynamics approach's numerical benefits by avoiding the mass matrix's inversion. 
Throughout the paper, we do not utilize the analytical constrained dynamics. However, we include it here to illustrate the standard derivation of the ECBF and highlight our equivalent approach, A-ECBF, that requires no additional matrix operations. 
\subsubsection{Review of ECBFs}
Consider the state $x=\begin{bmatrix} q^T, \dot{q}^T \end{bmatrix}^T \in D \subset TQ$, where $D$ is an open set of admissible states. By solving analytically the constrained dynamics we can obtain the following state-space representation,
\begin{align}
    \dot{x} = f(x) + g(x) u
    \label{eq:affine} 
\end{align}
where, $f(x) \in \mathbb{R}^{2n}, g(x) \in \mathbb{R}^{2n \times m}$ and  $u \in \mathbb{R}^m$. 
A safe region can be defined by,
\begin{align}
    \mathcal{C} = \{ x \in D : h(x) \geq 0 \},
\end{align}
where $h: D \rightarrow \mathbb{R}$ is a smooth function with relative degree $r_b$, i.e.,
\begin{align}
    h^{(r_b)}(x, u) = L_f^{r_b} h(x) + L_g L_f^{r_b-1} h(x) u. 
    \label{eq:cbf_reldeg}
\end{align}
 where, $L_f^{r_b} h(x), \hspace{0.5em} L_g L_f^{r_b-1}h(x) \in \mathbb{R}$ are the respective Lie derivatives of $h(x)$ and $L_g L_f^{r-1}h(x)$ is assumed to be non-zero.
We stack the lower derivatives to form the following state,
\begin{align}
    \eta_b(x) = \begin{bmatrix} h(x) \\ h^{(1)}(x) \\    \vdots \\ h^{(r_b-1)}(x) \end{bmatrix} = \begin{bmatrix} h(x) \\ L_f h(x) \\    \vdots \\ L^{r_b-1}_f h(x) \end{bmatrix}
\end{align}
and construct a linear representation of the dynamics of $h(x)$ through the mapping $L_f^{r_b} h(x) + L_g L_f^{r_b-1} h(x) u = \nu_b$, 
\begin{align}
    \dot{\eta}_b &= \underbrace{\begin{bmatrix} \textbf{0} & I \\ \textbf{0} & \textbf{0} \end{bmatrix}}_{F_b} \eta_b + \underbrace{\begin{bmatrix} \textbf{0} \\ I \end{bmatrix}}_{G_b} \nu_b 
    \label{eq:hlinearsys} \\
    h(x) &= \underbrace{\begin{bmatrix} 1 & 0 & 0 & ... & 0 \end{bmatrix}}_{C_b} \eta_b 
    \label{eq:houtput}
\end{align}
Next, we need a constraint on $\nu_b$ such that $h(x) \geq 0$. Applying the feedback $\nu_b = -K_{\alpha} \eta_b$, the trajectory becomes $h(x) = C_b e^{(F_b - G_b K_{\alpha})t} \eta_b (x_0)$. 

By the comparison lemma, setting $\nu_b \geq - K_{\alpha} \eta_b $ implies $h(x(t)) \geq  C_b e^{(F_b - G_b K_{\alpha})t} \eta_b (x_0)$. 
\begin{definition}[\bfseries Exponential Control Barrier Function~\cite{ames2019control}]
Given a set $\mathcal{C} \subset D$ defined as the super-level set of a $r_b$ times continuously differentiable function $h: D \rightarrow \mathbb{R} $, then $h$ is an exponential control barrier function (ECBF) if there exists $K_{\alpha} \in \mathbb{R}^{1 \times r_b}$ such that, 
\begin{align}
    \sup_{u \in U} \left[L^{r_b}_f h(x) + L_g L^{r_b-1}_f h(x) u \right] \geq -K_{\alpha} \eta_b(x)
    \label{eq:ECBFdef}
\end{align}
$\forall x \in \text{Int}(\mathcal{C})$ originates $h(x(t)) \geq C_b e^{(F_b - G_b K_{\alpha}t)} \eta_b (x_0) \geq 0$ for $h(x_0) \geq 0$.
\end{definition}

The closed loop system $\dot{\eta}_b = (F_b - G_b K_{\alpha}) \eta_b$ has a number of $r_b$ roots that are dependent on the selection of $K_{\alpha}$ and denoted by $p_b = -[p_1, ..., p_{r_b}]$. We will construct a family of functions $\mathsf{B}_i:D \rightarrow \mathbb{R}$ such that,
\begin{align}
    \mathsf{B}_0(x) &= h(x) \\
    \mathsf{B}_{i}(x) &= \dot{\mathsf{B}}_{i-1}(x) + p_{i} \mathsf{B}_{i-1}(x), \hspace{1em} \forall i = 1, ... , r_b
\end{align}
Note that if we choose $p_i$ such that $\mathsf{B}_i(x) \geq 0$ and $p_i > 0$, then $\mathsf{B}_{i-1}(x) \geq 0$. However, we only need to guarantee this once,  thus $\mathsf{B}_{i-1}(x_0) \geq 0$ is enough. This is the basis for the next theorem that is proved in~\cite{ames2019control}.

\begin{theorem}
\cite{ames2019control} Consider the closed loop system  $\dot{\eta}_b = (F_b - G_b K_{\alpha}) \eta_b$. If $K_{\alpha}$ is chosen to enforce its roots $p_b = -[p_1, ..., p_{r_b}]$ to be,
\begin{align}
    p_i > 0, \hspace{0.5cm}
    p_i \geq - \frac{\dot{\mathsf{B}}_{i-1}(x_0)}{\mathsf{B}_{i-1}(x_0)},
    \hspace{0.5cm} \forall i = 1,...,r_b
\end{align}
Then, the constraint $\nu_b \geq -K_{\alpha} \eta_b$ applied on \eqref{eq:hlinearsys}, renders $h(x)$ an ECBF.
\end{theorem}
\subsubsection{Formulation of A-ECBFs}
In our inverse dynamics formulation, the decision variable $X$ includes the acceleration $\ddot{q}$ and the control output $u$, which are related through the system dynamics. We can avoid solving the constrained dynamics by using the accelerations $\ddot{q}$ instead of $u$, and the quadratic program can implicitly solve their relationship. We can obtain an alternative expression for the $r_b$ derivative of $h(x)$ as,
\begin{align}
    h^{(r_b)}(x, \ddot{q}) = L^{r_b}_F h(x) + L_G L^{r_b-1}_F h(x) \ddot{q}
    \label{eq:CBF_eq}
\end{align}
where, $F$ and $G$ are found by choosing the coordinates $\eta = \begin{bmatrix} h(x), h^{(1)}(x), ..., h^{(r_b-1)}(x) \end{bmatrix}^\top$ and constructing the following linear system,
\begin{align}
    \dot{\eta} = \underbrace{\begin{bmatrix} \textbf{0} & I \\ \textbf{0} & \textbf{0} \end{bmatrix}}_{F} \eta + \underbrace{\begin{bmatrix} \textbf{0} \\ I \end{bmatrix}}_{G} \ddot{q}
\end{align}

Using this expression, we present an equivalent definition of \eqref{eq:ECBFdef}, considering joint accelerations and avoiding the explicit solution of the constrained dynamics needed for \eqref{eq:affine}.

\begin{definition}[\bfseries Acceleration based exponential control barrier function (A-ECBF)]
The function $h(x)$ is an A-ECBF if there exists $K_{\alpha}$ such that,
\begin{align}
    \sup_{\ddot{q} \in \mathbb{R}^n} \left[ L^{r_b}_F h(x) + L_G L^{r_b-1}_F h(x) \ddot{q} \right] \geq -K_{\alpha} \eta_b(q,\dot{q})
    \label{eq:AECBF}
\end{align}
$\forall x \in \text{Int}(\mathcal{C})$ originates $h(x(t)) \geq C_b e^{(F_b - G_b K_{\alpha}t)} \eta_b (x_0) \geq 0$ for $h(x_0) \geq 0$.
\end{definition}

The design of $K_{\alpha}$ follows the \textbf{Theorem 1} rationale. Once we have a suitable value for it, we include the constraint \eqref{eq:AECBF} into our inverse dynamics controller.
In other words, we simply add the A-ECBF certificate as an additional inequality constraint as shown below.
\begin{align*}
    X^* = \argminB_{X \in \mathcal{X}} &|| J_y(q) \ddot{q} + \dot{J}_y(q,\dot{q})\dot{q} - y^*||^2 +  X^T \Gamma X \\
    \st & D_{eq}(q) X = -C(q,\dot{q}) - G(q) \\
    & C_{eq}(q) X = -J(q) \dot{q} \\
    & L^{r_b}_F h(q, \dot{q}) + L_G L^{r_b-1}_F h(q, \dot{q}) \ddot{q} \geq -K_{\alpha} \eta_b(q, \dot{q})
\end{align*}
This controller enforces forward invariance of the set $\mathcal{C}$ using our equivalent definition of an A-ECBF in the inverse dynamics formula. 
In systems that start from rest, i.e, $x(t_0) = \dot{x}(t_0) = 0$ we note that \textbf{Theorem 1} simplifies to $p_i > 0, \forall i = 1,...,r_b$. For second-order systems, it means that $K_{\alpha}$ accepts solutions that form critically damped or over-damped oscillations. It makes the solutions of $h(x)$ to avoid zero crossings and maintain $h(x) \geq 0$.

\begin{figure*}[t]
\vspace{2mm}
\centering
\includegraphics[width=0.85\linewidth]{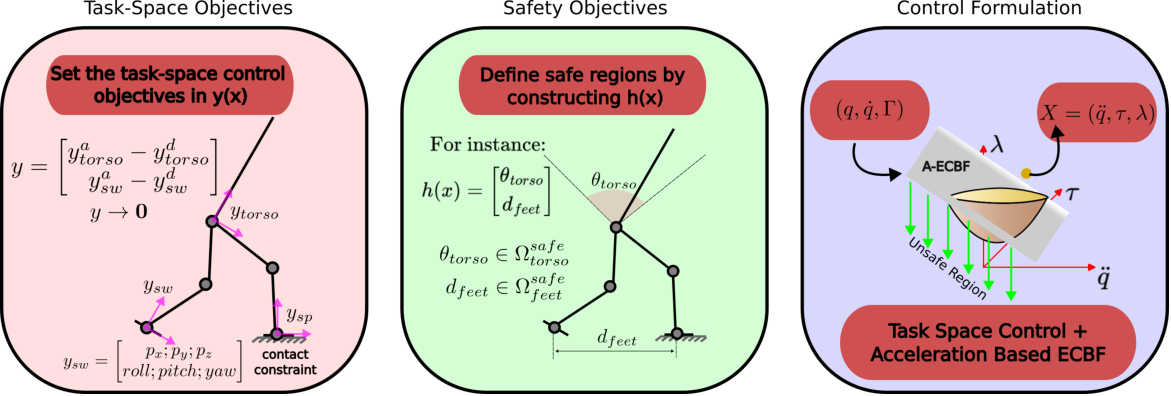}
\caption{The safe task-space control framework starts by defining control objectives and safety sets according to the application. The optimization will search a control action in the space of safe solutions.}
\vspace{-5mm}
\label{fig:framework}
\end{figure*}




%% file: sections/results.tex
\section{Simulation and Experimental Results}
\label{sec:results}
\begin{figure}[h]
\centering
\includegraphics[width=\columnwidth]{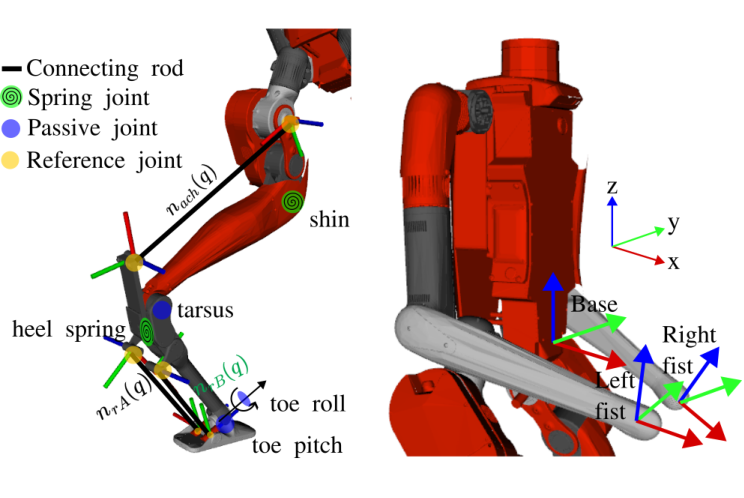}
\caption{The figure on the left shows the passive joints and the closed chain loops in the robot legs. The figure on the right shows the fist and the base frames.}
\label{fig:leg_joints}
\vspace{-4mm}
\end{figure}

We implement the controller in simulation and the robotic hardware using a unique code structure and controller gains. Our test bed is Digit, a 3D bipedal robot with arms, legs, and a torso developed by Agility Robotics. It weighs 45 Kg and has 30 joints with 20 motors. Each leg presents three closed kinematic chains and two spring joints, as seen in Fig. \ref{fig:leg_joints}.

We consider the springs as rigid joints to simplify the control problem because of their high stiffness. Those spring joints will be considered part of the kinematic constraints (kc). The entire kinematic constraints are defined as $n_{kc} = \begin{bmatrix} n^L_{kc} & n^R_{kc} \end{bmatrix}^T$, where $L/R$ stands for left and right, respectively, and
\begin{align}
    n^{L/R}_{kc} = \begin{bmatrix} n_{ach} & n_{rA} & n_{rB} & q_{shin} & q_{heel-spring} \end{bmatrix}_{L/R}^T
\end{align}
where, $n_{ach}$ is the length of the achilles rod, $n_{rA}$ and $n_{rB}$ are the lengths of the rods connecting to the ankles while $q_{shin}$ and $q_{heel-spring}$ are the spring joints considered as fixed, as shown in Fig. \ref{fig:leg_joints}.
We parameterize time by $\tau = t - t_0$, where $t_0$ indicates the starting time of a step. We show the results in both, plots and video~\footnote{https://youtu.be/vNTIcODR6cI}, for different whole body motions including walking. 



\subsection{Task: Squatting and bowing}
In this case we showcase the whole-body controller without safety constraints. During the Double Support domain, the controlled outputs are:
\begin{align}
    y(q,t) = \begin{bmatrix} p_{CoM}(q) \\ \theta_{torso}(q) \\ q_{arms}(q) \end{bmatrix} 
    - \begin{bmatrix} p^d_{CoM}(t) \\ \theta^d_{torso}(t) \\ q^d_{arms}(t) \end{bmatrix} 
    \label{eq: yout_euler}
\end{align}
where, $p_{CoM} \in \mathbb{R}^3$ is the position of the center of mass, $\theta_{torso}  \in \mathbb{R}^3$ is the torso orientation expressed in Euler ZYX angles and, $q_{arms}  \in \mathbb{R}^8$ are the angular positions of the joints on the left and right arm. Note that during this domain, the robot must keep both feet in contact with the ground, i.e., meet the ZMP and friction constraints.

\begin{figure}[t]
\centering
\includegraphics[width=\columnwidth]{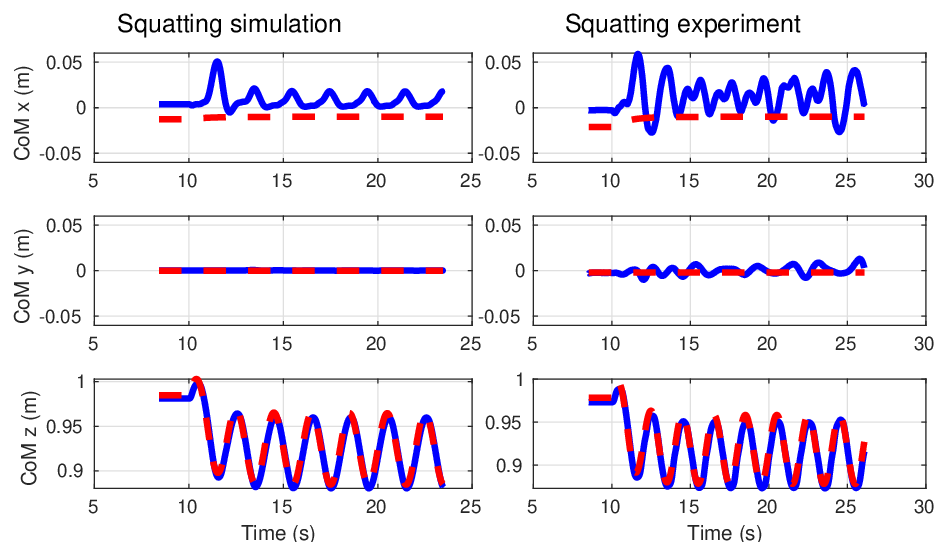}
\caption{The squatting motion starts at $t=10s$ for both simulation and experiment. The blue lines are the actual outputs, and the red dashed lines represent the reference trajectories.}
\label{fig:squatting}
\vspace{-4mm}
\end{figure}

We test two continuous actions:  (1) a squatting motion by specifying a sinusoidal reference to the CoM's height and (2) a bowing motion by commanding the torso pitch to extend and return to its initial pose.  
The squatting reference is,
\begin{align}
    p^d_{CoM}(t) =
    \begin{bmatrix} 
        -0.02 \\ 0 \\ 1 - 0.12(1 - e^{-\tau}) + 0.03 sin(\pi \tau)
    \end{bmatrix}
\end{align}
During this motion, we keep the torso orientation straight $\theta^d_{torso} = \textbf{0}$ and the arms fixed. Fig. \ref{fig:squatting} shows the controller's performance in simulation and hardware experiments. 

\begin{figure}[h]
\centering
\vspace{1mm}
\includegraphics[width=\columnwidth]{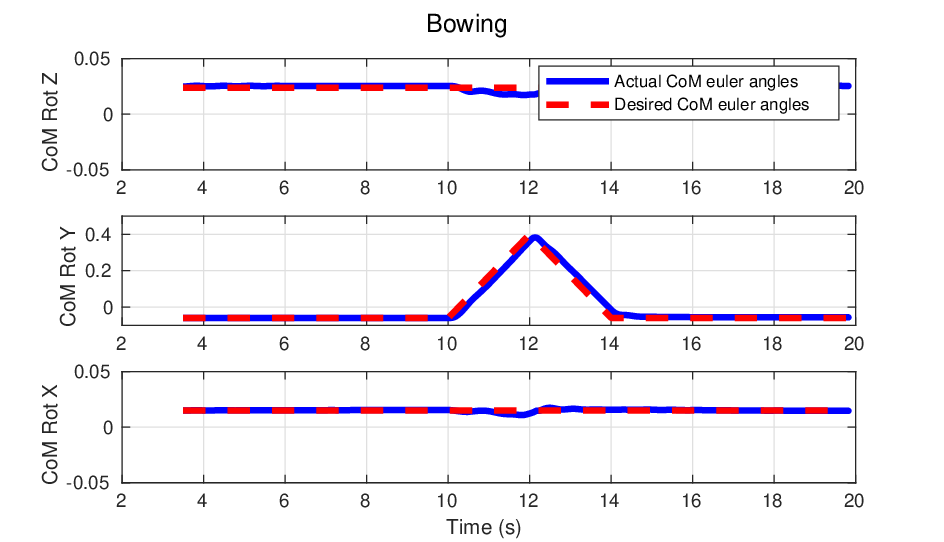}
\caption{Tracking the torso orientation while realizing a bowing movement between 10-14 seconds.}
\label{fig:bowing}
\vspace{-4mm}
\end{figure}

Regarding the bowing motion, we set the following reference for the torso pitch,
\begin{align}
    \theta^d_{torso}(t) = \begin{bmatrix} 0 & 0.45 \max(3-|\tau-3|, 0) & 0 \end{bmatrix}^\top
\end{align}
and the other tasks are specified as $p^d_{CoM} = \begin{bmatrix} 0 & 0 & 0.95 \end{bmatrix}^T$ and the arms fixed. Applying our controller results on the behavior observed in Fig. \ref{fig:bowing}, for the hardware experiment.

\subsection{Task: Arm motion with height limits}
\begin{figure}[h]
\centering
\vspace{2mm}
\includegraphics[width=\columnwidth]{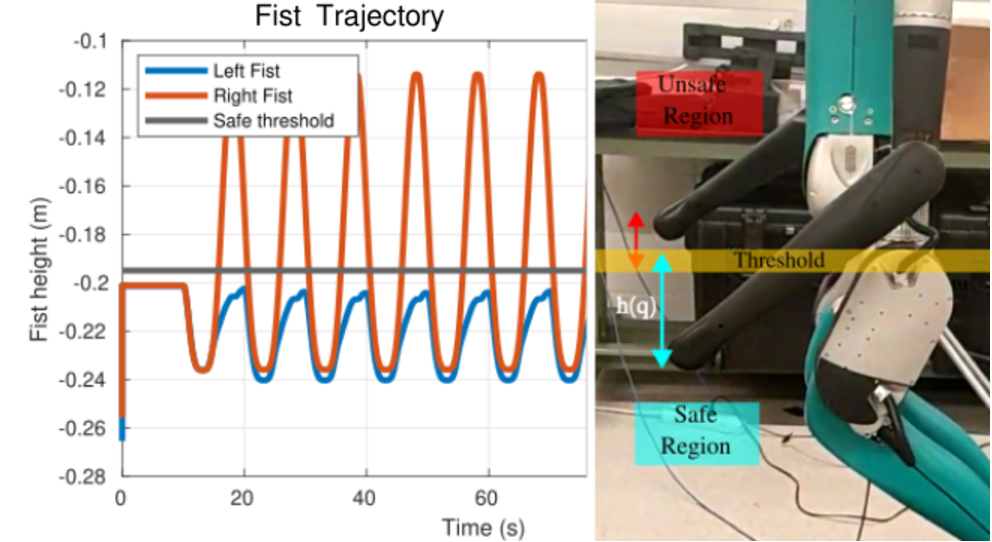}
\caption{The trajectory of the left fist and right fist. The left fist is constrained by an A-ECBF that prevents it from crossing the safe threshold, while the right fist is not constrained and crosses it at $t=30s$ during the hardware experiments.}
\label{fig:arms}
\end{figure}

\begin{figure}[h]
\centering
\includegraphics[width=\columnwidth]{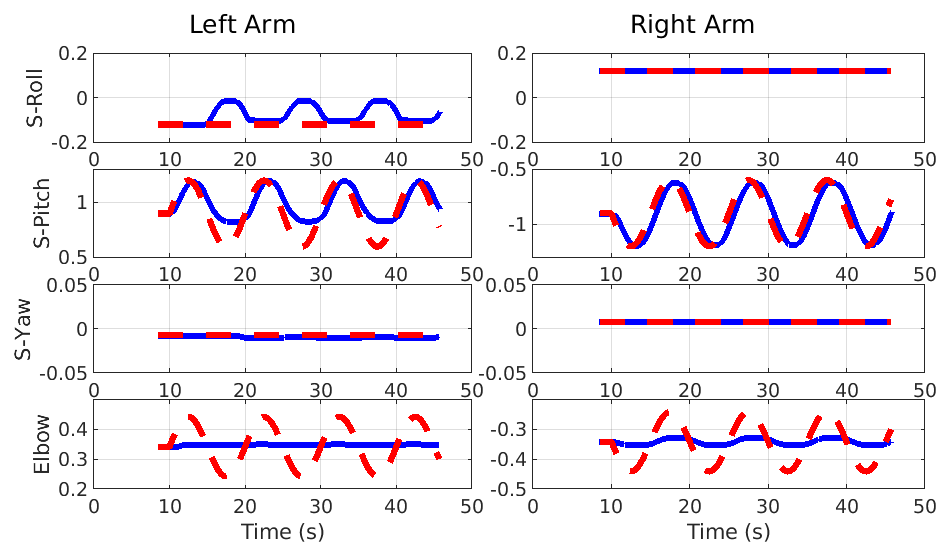}
\caption{Arm joint tracking during the hardware experiment. The blue lines represent the actual joint positions, and the red dashed lines show the desired trajectories.}
\label{fig:Otherarms}
\vspace{-4mm}
\end{figure}

We command an arm trajectory during the Double Support domain, keeping the robot's center of mass and torso orientation fixed and straight. We will focus on the safety-critical feature of the controller by providing an A-ECBF for the fist height with respect to the base frame, as seen in Fig. \ref{fig:leg_joints}. To illustrate the effect of the A-ECBF, we will command both arms, left and right, with the same reference but only equip the left fist with the safety certificate. We consider an imaginary safe set, defined as,
\begin{align}
    h(q) = -p^{L}_z(q) - 0.195 \geq 0
\end{align}
where, $p^L_z(q)$ is the z-position of the fist of the left arm w.r.t to the torso base, as seen in Fig. \ref{fig:arms}. Since $h(x)$ is relative degree two, the A-ECBF constraint takes the form of $J_{p_z} \ddot{q}  + \dot{J}_{p_z} \dot{q} \geq - K_{\alpha} \eta$. Moreover,
\begin{align}
    \dot{\eta} = \begin{bmatrix}
        0 & 1 \\ 0 & 0
    \end{bmatrix} \eta + 
    \begin{bmatrix}
        0 \\ 1
    \end{bmatrix} \ddot{q}
\end{align}
where, $J_{p_z} = -\frac{\partial p_z^L}{\partial q}$ and $\eta = [-p^{L}_z - 0.195, -\dot{p}^L_z]^\top$. We design $K_{\alpha}$ such that the roots of $\dot{\eta} = (F-GK_{\alpha}) \eta$ follow \textbf{Theorem 1}.
The output references for both left and right arm are $q^d_{arm} = \begin{bmatrix} q^d_{arm_L}, q^d_{arm_R} \end{bmatrix}^T$ with,
\begin{align}
    q^d_{arm_L} = \begin{bmatrix} 0 & 0.3 \sin(\frac{\pi}{5} \tau) & 0 & 0.2 \sin(\frac{\pi}{5} \tau) \end{bmatrix}^\top, \hspace{1em} \\
    q^d_{arm_R} = \begin{bmatrix} 0 & -0.3 \sin(\frac{\pi}{5} \tau) & 0 & -0.2 \sin(\frac{\pi}{5} \tau) \end{bmatrix}^\top
\end{align}
Applying the QP-based controller with the A-ECBF results on the fist motions shown in Fig. \ref{fig:arms} in hardware. We note that the right fist violates the safe zone to reach its desired target, while the left fist avoids those trajectories but will try to reach the other objectives. Fig. \ref{fig:Otherarms} shows the tracking of the other objectives. 

\subsection{Task: Walking with the A-LIP template model}
\begin{figure}[h]
\centering
\includegraphics[width=\columnwidth]{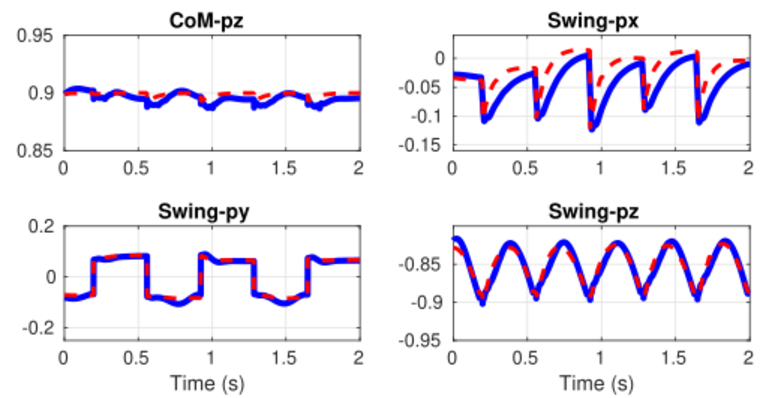}
\caption{Task space tracking under the ALIP planner in the Mujoco simulator. We note a good tracking performance during the walking gait, and a similar performance is observed for the orientation tasks.  }
\label{fig:yALIP_tracking}
\vspace{-4mm}
\end{figure}

In the following tasks, we use a single support domain that will enable bipedal walking. We use the Angular Momentum-based Linear Inverted Pendulum (ALIP) model to generate stable walking patterns~\cite{gong2020angular}. In the ALIP, the support ankle (pitch and roll) is rendered passive to predict the angular momentum at the end of the step. This strategy allows us to plan for stabilizing foot positioning targets $(u_x, u_y)$ and the feet height profile to impact at a specified period $T$. A complete description of the planner can be found at \cite{gong2020angular, gibson2021terrain}. The swing foot outputs are defined as:
\begin{align}
    p^d_{swing}(\tau) = \begin{bmatrix} 
    x_{sw}(\tau_0) + s (u_x - x_{sw}(\tau_0)) \\
    y_{sw}(\tau_0) + s (u_y - y_{sw}(\tau_0)) \\
    0.08 sin(\frac{2\pi \tau}{T})
    \end{bmatrix}
\end{align}
where $s = (1 - e^{-5 \frac{\tau}{T}})$ is an smoothing factor.
The other ALIP requirements are: 
the CoM height is constant ($H = 0.9$), the orientation of the torso is kept vertical, and the swing foot remains flat. Hence, the actual and desired outputs are defined as:
\begin{align}
    y_{ALIP}^a = \begin{bmatrix} z_{CoM} \\ \theta_{torso} \\ p_{swing} \\ \theta_{swing} \end{bmatrix}, 
    y_{ALIP}^d = \begin{bmatrix} H \\ \textbf{0}_{3 \times 1} \\ p^d_{swing}(\tau) \\ \textbf{0}_{3 \times 1} \end{bmatrix},
    \label{eq: yALIP}
\end{align}
The controller's application over the outputs defined in \eqref{eq: yALIP} produces a stable gait with an average forward speed of 0.2 m/s. The task space tracking for the position of the outputs is summarized in \figref{fig:yALIP_tracking}.

\subsection{Task: Collision avoidance}
Continuing with the ALIP based gait, we show the effect of an instantaneous ($\delta t = 0.15s$) lateral external force at the torso with magnitude $F_{ext} =  -30N$. This produces a leg motion that leads to the collision between the legs, as seen in Fig. \ref{fig:ALIP_collision}. However, by providing collision safety through an A-ECBF, we can avoid this event and recover the balance. The safe set is defined by,
\begin{align}
    h_{ALIP}(q) = y_{sw}(q) - 0.07 \geq 0
\end{align}
where, $y_{sw}(q)$ is the y-position of the swing foot w.r.t the support foot. By enforcing the constraint, the swing foot will maintain a distance margin that will avoid collisions. Fig. \ref{fig:ALIP_collision} shows that the reference provided by the ALIP planner crosses the safety region, but the A-ECBF disallows the foot to enter that region.

%% file: sections/conclusions.tex
\section{Conclusions}
\label{sec:conclusion}
The whole body controller presented realizes stable motions that respect its closed chain kinematic, ZMP, and other physical constraints. We achieved fast squatting and bowing movements that show the controller's capabilities during double support. We also conducted walking experiments with 0.2m/s of speed and a stepping time of $T=0.35s$ to show fast single support events handling. Furthermore, we showed the formulation of the A-ECBF to provide control safety. This safe controller has the numerical benefit of expressing the dynamics and the constraints separately and avoiding computation of the constrained dynamics. The results of the A-ECBF were applied to both the arms and the legs to show its effectiveness in different scenarios. In general, the results show similar performance between simulation and hardware experiments regarding tracking and safety.

\begin{figure}[!t]
\centering
\includegraphics[width=\columnwidth]{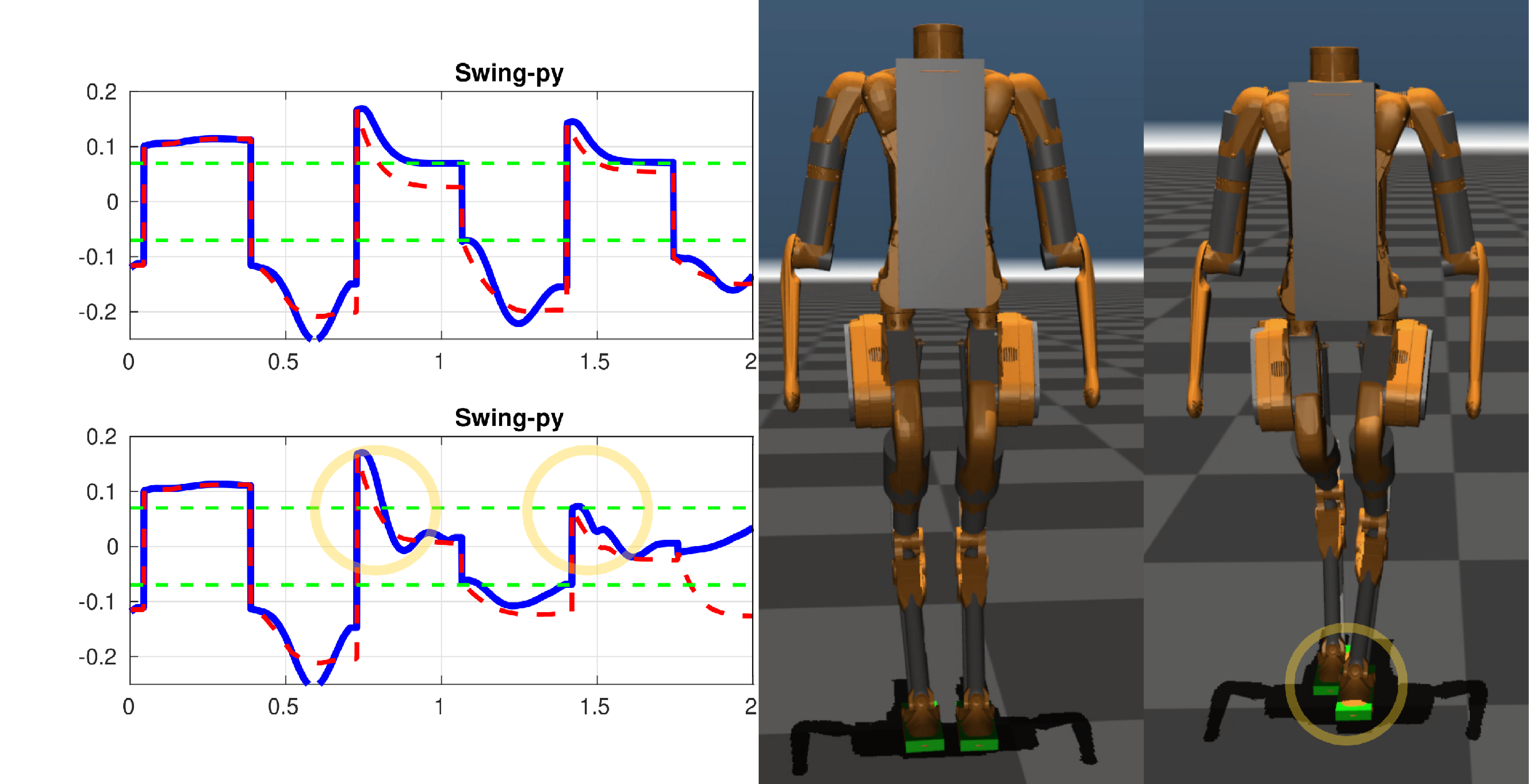}
\caption{During the Mujoco simulation, we observe the swing foot tracking in the y-direction. The upper plot has an A-ECBF, while the lower does not. The foot placement reference  (red dashed line), crosses the safe region in both plots. However, the A-ECBF prevents the foot from entering it. The yellow circles show the collision events. }
\label{fig:ALIP_collision}
\vspace{-4mm}
\end{figure}



%% file: root.bbl
\begin{thebibliography}{10}
\providecommand{\url}[1]{#1}
\csname url@rmstyle\endcsname
\providecommand{\newblock}{\relax}
\providecommand{\bibinfo}[2]{#2}
\providecommand\BIBentrySTDinterwordspacing{\spaceskip=0pt\relax}
\providecommand\BIBentryALTinterwordstretchfactor{4}
\providecommand\BIBentryALTinterwordspacing{\spaceskip=\fontdimen2\font plus
\BIBentryALTinterwordstretchfactor\fontdimen3\font minus \fontdimen4\font\relax}
\providecommand\BIBforeignlanguage[2]{{%
\expandafter\ifx\csname l@#1\endcsname\relax
\typeout{** WARNING: IEEEtran.bst: No hyphenation pattern has been}%
\typeout{** loaded for the language `#1'. Using the pattern for}%
\typeout{** the default language instead.}%
\else
\language=\csname l@#1\endcsname
\fi
#2}}

\bibitem{prajna2006barrier}
S.~Prajna, ``Barrier certificates for nonlinear model validation,'' \emph{Automatica}, vol.~42, no.~1, pp. 117--126, 2006.

\bibitem{prajna2004safety}
S.~Prajna and A.~Jadbabaie, ``Safety verification of hybrid systems using barrier certificates,'' in \emph{International Workshop on Hybrid Systems: Computation and Control}.\hskip 1em plus 0.5em minus 0.4em\relax Springer, 2004, pp. 477--492.

\bibitem{hsu2015control}
S.~Hsu, X.~Xu, and A.~D. Ames, ``Control barrier function based quadratic programs with application to bipedal robotic walking,'' in \emph{American Control Conference, {ACC} 2015, Chicago, IL, USA, July 1-3, 2015}, IEEE.\hskip 1em plus 0.5em minus 0.4em\relax {IEEE}, 2015, pp. 4542--4548.

\bibitem{nguyen2016exponential}
Q.~Nguyen and K.~Sreenath, ``Exponential control barrier functions for enforcing high relative-degree safety-critical constraints,'' in \emph{2016 American Control Conference (ACC)}.\hskip 1em plus 0.5em minus 0.4em\relax IEEE, 2016, pp. 322--328.

\bibitem{ames2019control}
A.~D. Ames, S.~Coogan, M.~Egerstedt, G.~Notomista, K.~Sreenath, and P.~Tabuada, ``Control barrier functions: Theory and applications,'' in \emph{2019 18th European Control Conference (ECC)}.\hskip 1em plus 0.5em minus 0.4em\relax IEEE, 2019, pp. 3420--3431.

\bibitem{del2017joint}
A.~Del~Prete, ``Joint position and velocity bounds in discrete-time acceleration/torque control of robot manipulators,'' \emph{IEEE Robotics and Automation Letters}, vol.~3, no.~1, pp. 281--288, 2017.

\bibitem{khazoom2022humanoid}
C.~Khazoom, D.~Gonzalez-Diaz, Y.~Ding, and S.~Kim, ``Humanoid self-collision avoidance using whole-body control with control barrier functions,'' in \emph{2022 IEEE-RAS 21st International Conference on Humanoid Robots (Humanoids)}.\hskip 1em plus 0.5em minus 0.4em\relax IEEE, 2022, pp. 558--565.

\bibitem{nakanishi2008operational}
J.~Nakanishi, R.~Cory, M.~Mistry, J.~Peters, and S.~Schaal, ``Operational space control: A theoretical and empirical comparison,'' \emph{The International Journal of Robotics Research}, vol.~27, no.~6, pp. 737--757, 2008.

\bibitem{paredes2022resolved}
V.~Paredes and A.~Hereid, ``Resolved motion control for 3d underactuated bipedal walking using linear inverted pendulum dynamics and neural adaptation,'' \emph{arXiv preprint arXiv:2208.01786}, 2022.

\bibitem{nakanishi2007inverse}
J.~Nakanishi, M.~Mistry, and S.~Schaal, ``Inverse dynamics control with floating base and constraints,'' in \emph{Proceedings 2007 IEEE International Conference on Robotics and Automation}.\hskip 1em plus 0.5em minus 0.4em\relax IEEE, 2007, pp. 1942--1947.

\bibitem{mistry2010inverse}
M.~Mistry, J.~Buchli, and S.~Schaal, ``Inverse dynamics control of floating base systems using orthogonal decomposition,'' in \emph{2010 IEEE international conference on robotics and automation}.\hskip 1em plus 0.5em minus 0.4em\relax IEEE, 2010, pp. 3406--3412.

\bibitem{herzog2014balancing}
A.~Herzog, L.~Righetti, F.~Grimminger, P.~Pastor, and S.~Schaal, ``Balancing experiments on a torque-controlled humanoid with hierarchical inverse dynamics,'' in \emph{Proc. IEEE/RSJ Int. Conf. Intelligent Robots and Systems}, Sept. 2014, pp. 981--988.

\bibitem{reher2020inverse}
J.~Reher, C.~Kann, and A.~D. Ames, ``An inverse dynamics approach to control lyapunov functions,'' in \emph{2020 American Control Conference (ACC)}.\hskip 1em plus 0.5em minus 0.4em\relax IEEE, 2020, pp. 2444--2451.

\bibitem{nguyen20163d}
Q.~Nguyen, A.~Hereid, K.~Sreenath, J.~W. Grizzle, and A.~D. Ames, ``{3D} dynamic walking on stepping stones with control barrier functions,'' in \emph{Proc. IEEE 55\textsuperscript{th} Conf. Decision and Control (CDC)}.\hskip 1em plus 0.5em minus 0.4em\relax Las Vegas, NV: IEEE, Dec. 2016, pp. 827--834.

\bibitem{nguyen2015safety}
Q.~Nguyen and K.~Sreenath, ``Safety-critical control for dynamical bipedal walking with precise footstep placement,'' \emph{IFAC-PapersOnLine}, vol.~48, no.~27, pp. 147--154, 2015.

\bibitem{grizzle2014models}
J.~W. Grizzle, C.~Chevallereau, R.~W. Sinnet, and A.~D. Ames, ``Models, feedback control, and open problems of {3D} bipedal robotic walking,'' \emph{Automatica}, vol.~50, no.~8, pp. 1955--1988, 2014.

\bibitem{moro2018whole}
F.~L. Moro and L.~Sentis, ``Whole-body control of humanoid robots,'' \emph{Humanoid Robotics: A Reference}, Jan. 2018.

\bibitem{reher2021control}
J.~Reher and A.~D. Ames, ``Control lyapunov functions for compliant hybrid zero dynamic walking,'' \emph{arXiv preprint arXiv:2107.04241}, 2021.

\bibitem{vukobratovic2004zero}
M.~Vukobratovi{\'c} and B.~Borovac, ``Zero-moment point: thirty five years of its life,'' \emph{International Journal of Humanoid Robotics}, vol.~1, no.~01, pp. 157--173, 2004.

\bibitem{gong2020angular}
Y.~Gong and J.~W. Grizzle, ``{Zero Dynamics, Pendulum Models, and Angular Momentum in Feedback Control of Bipedal Locomotion},'' \emph{Journal of Dynamic Systems, Measurement, and Control}, vol. 144, no.~12, 10 2022, 121006.

\bibitem{gibson2021terrain}
G.~Gibson, O.~Dosunmu-Ogunbi, Y.~Gong, and J.~Grizzle, ``Terrain-adaptive, alip-based bipedal locomotion controller via model predictive control and virtual constraints,'' in \emph{2022 IEEE/RSJ International Conference on Intelligent Robots and Systems (IROS)}, 2022, pp. 6724--6731.

\end{thebibliography}
